\documentclass[10pt,twocolumn,letterpaper]{article}

\usepackage{cvpr}
\usepackage{times}
\usepackage{epsfig}
\usepackage{graphicx}
\usepackage{amsmath}
\usepackage{amssymb}

\usepackage{blindtext}
\usepackage{bbm}
\usepackage{enumitem}
\usepackage{epsfig}
\usepackage[ruled]{algorithm2e}
\usepackage{algorithmic}
\usepackage{multirow}
\usepackage{color}
\usepackage{xcolor}
\usepackage{color, colortbl}
\usepackage{array}
\usepackage[english]{babel} % English language/hyphenation
\usepackage{amsmath,amsfonts,amsthm} % Math packages
\usepackage{mathtools}
\usepackage{pifont}
\usepackage{authblk}

\expandafter\def\expandafter\normalsize\expandafter{%
	\normalsize\setlength\abovedisplayskip{3pt}}
\expandafter\def\expandafter\normalsize\expandafter{%
	\normalsize\setlength\belowdisplayskip{3pt}}
\usepackage[pagebackref=true,breaklinks=true,letterpaper=true,colorlinks,bookmarks=false]{hyperref}

\cvprfinalcopy % *** Uncomment this line for the final submission

\def\cvprPaperID{188} % *** Enter the CVPR Paper ID here

\ifcvprfinal\fi
\begin{document}

\title{Auto-Encoding Twin-Bottleneck Hashing}
\author[1]{Yuming Shen\thanks{Equal Contribution}
}
\author[1]{Jie Qin\footnote[1]\thanks{Corresponding Author}
}
\author[1]{Jiaxin Chen\footnote[1]
}
\author[1]{Mengyang Yu
}

\author[1]{\\Li Liu
}
\author[1]{Fan Zhu
}
\author[2]{Fumin Shen
}
\author[1]{Ling Shao
}
\vspace{-4ex}
\affil[1]{Inception Institute of Artificial Intelligence (IIAI), Abu Dhabi, UAE}
\affil[2]{Center for Future Media, University of Electronic Science and Technology of China, China}
\affil[ ]{\small \texttt{ymcidence@gmail.com}}

\maketitle
\thispagestyle{empty}

%%%%%%%%% ABSTRACT
\begin{abstract}
	Conventional unsupervised hashing methods usually take advantage of similarity graphs, which are either pre-computed in the high-dimensional space or obtained from random anchor points. On the one hand, existing methods uncouple the procedures of hash function learning and graph construction. On the other hand, graphs empirically built upon original data could introduce biased prior knowledge of data relevance, leading to sub-optimal retrieval performance. In this paper, we tackle the above problems by proposing an efficient and adaptive code-driven graph, which is updated by decoding in the context of an auto-encoder. Specifically, we introduce into our framework twin bottlenecks (i.e., latent variables) that exchange crucial information collaboratively. One bottleneck (i.e., binary codes) conveys the high-level intrinsic data structure captured by the code-driven graph to the other (i.e., continuous variables for low-level detail information), which in turn propagates the updated network feedback for the encoder to learn more discriminative binary codes. The auto-encoding learning objective literally rewards the code-driven graph to learn an optimal encoder. Moreover, the proposed model can be simply optimized by gradient descent without violating the binary constraints. Experiments on benchmarked datasets clearly show the superiority of our framework over the state-of-the-art hashing methods. Our source code can be found at \url{https://github.com/ymcidence/TBH}.
\end{abstract}

%%%%%%%%% BODY TEXT
\vspace{-3ex}\section{Introduction}\label{sec_1}
Approximate Nearest Neighbour (ANN) search has attracted ever-increasing attention in the era of big data. Thanks to the extremely low costs for computing Hamming distances, binary coding/hashing has been appreciated as an efficient solution to ANN search. Similar to other feature learning schemes, hashing techniques can be typically subdivided into supervised and unsupervised ones.
% Supervised hashing~\cite{ccaitq,bre,cnnh,dnnh,gcnh} typically utilizes semantic labels or pair-wise supervision during training. As a result, it largely favours the existing retrieval evaluation metric, which 
Supervised hashing~\cite{ccaitq,bre,dnnh,jmlh,cnnh,gcnh}, which highly depends on labels, is not always preferable since large-scale data annotations are unaffordable. Conversely, unsupervised hashing~\cite{itq,knnh,he2013k,isoh,seh,yu2016structure,distill}, provides a cost-effective solution for more practical applications. To exploit data similarities, existing unsupervised hashing methods~\cite{dgh,agh,sh,shen2018unsupervised,greedyhash} have extensively employed graph-based paradigms.
%Nevertheless, similarity graphs are usually expensively computed in the original high-dimensional Euclidean space, thus not scalable to large-scale data. In addition, once a graph is explicitly pre-computed, it cannot be updated adaptively, 
Nevertheless, existing methods usually suffer from the `\textbf{static graph}' problem. More concretely, they often adopt explicitly pre-computed graphs, introducing biased prior knowledge of data relevance. Besides, graphs cannot be adaptively updated to better model the data structure. The interaction between hash function learning and graph construction can be hardly established. The `static graph' problem greatly hinders the effectiveness of graph-based unsupervised hashing mechanisms.
%, though it generally underperforms supervised methods in similarity search.
%A variety of approaches have been proposed in this research scope, \eg, random projection~\cite{lsh,klsh}, alternative projection and quantization~\cite{itq}, graph-based paradigms~\cite{sph,dgh,agh}, and even deep learning based ones~\cite{dh,deepbit,sh}.
% In this paper, we focus on unsupervised hashing for more practical applications. 

\begin{figure}[t]
	\begin{center}
		\includegraphics[width=0.85\linewidth]{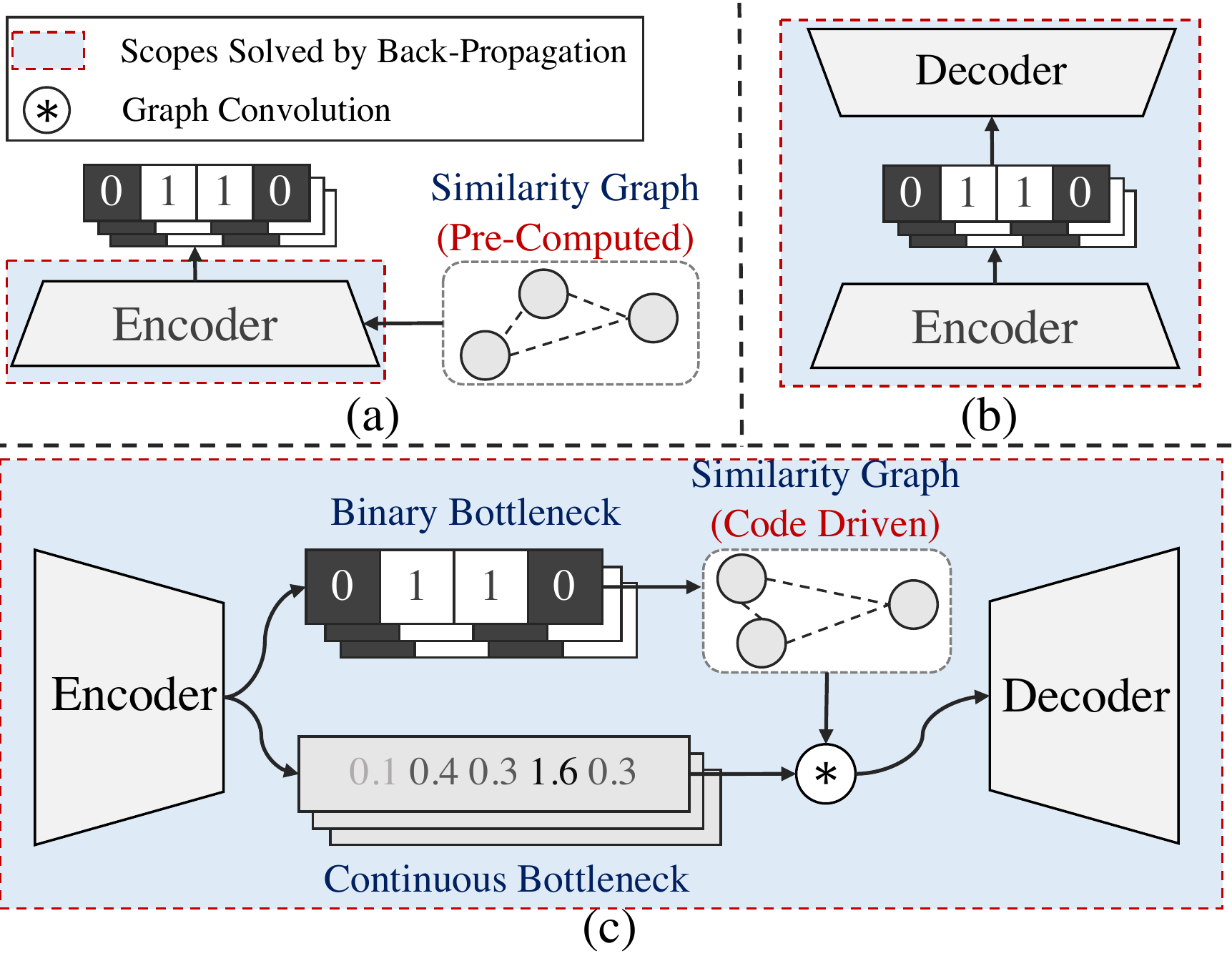}
	\end{center}\vspace{-1ex}
	%\vspace{-1ex}
	\caption{Illustration of the differences between \textbf{(a)} graph-based hashing, \textbf{(b)} unsupervised generative auto-encoding hashing, and \textbf{(c)} the proposed model (TBH).}\vspace{-2ex}
	\label{fig_1}
\end{figure}

In this work, we tackle the above long-standing challenge by proposing a novel adaptive graph, which is directly driven by the learned binary codes. The graph is then seamlessly embedded into a generative network that has recently been verified effective for learning reconstructive binary codes \cite{sgh,hashgan,bgan,bingan}. In general, our network can be regarded as a variant of Wasserstein Auto-Encoder \cite{wae} with two kinds of bottlenecks (\ie, latent variables). Hence, we call the proposed method Twin-Bottleneck Hashing (\textbf{TBH}). Fig.~\ref{fig_1} illustrates the differences between TBH and the related models. As shown in Fig.~\ref{fig_1} (c), the binary bottleneck (\textbf{BinBN}) contributes to constructing the code-driven similarity graph, while the continuous bottleneck (\textbf{ConBN}) mainly guarantees the reconstruction quality. Furthermore, Graph Convolutional Networks (GCNs)~\cite{gcn} are leveraged as a `tunnel' for the graph and the ConBN to fully exploit data relationships. As a result, similarity-preserving latent representations are fed into the decoder for high-quality reconstruction. Finally, as a reward, the updated network setting is back-propagated through the ConBN to the encoder, which can better fulfill our ultimate goal of binary coding.

More concretely, TBH tackles the `\textbf{static graph}' problem by directly leveraging the latent binary codes to adaptively capture the intrinsic data structure. To this end, an adaptive similarity graph is computed directly based on the Hamming distances between binary codes, and is used to guide the ConBN through neural graph convolution~\cite{hammond2011wavelets,gcn}.
% Different from \cite{dgh,agh,shen2018unsupervised,gcnh} where the graph is \textit{fixed}, our model guides the network to stochastically \textit{learn} the graph from the produced codes.
This design provides an optimal mechanism for efficient retrieval tasks by directly incorporating the Hamming distance into training. \emph{On the other hand}, as a side benefit of the twin-bottleneck module, TBH can also overcome another important limitation in generative hashing models~\cite{ba,sgh}, \ie, directly inputting the BinBN to the decoder leads to poor data reconstruction capability. For simplicity, we call this problem as `\textbf{deficient BinBN}'. Particularly, we address this problem by leveraging the ConBN, which is believed to have higher encoding capacity, for decoding. In this way, one can expect these continuous latent variables to preserve more entropy than binary ones. Consequently, the reconstruction procedure in the generative model becomes more effective.

In addition, during the optimization procedure, existing hashing methods often employ alternating iteration for auxiliary binary variables \cite{sdh} or even discard the binary constraints using some relaxation techniques \cite{dh}. In contrast, our model employs the distributional derivative estimator \cite{sgh} to compute the gradients across binary latent variables, ensuring that the binary constraints are not violated. Therefore, the whole TBH model can be conveniently optimized by the standard Stochastic Gradient Descent (SGD) algorithm.

% In addition, TBH follows the Wasserstein Auto-Encoder (WAE)~\cite{wae} to adversarially regularize latent variables for better code quality.

The main contributions of this work are summarized as:
\begin{itemize}
	\item We propose a novel unsupervised hashing framework by incorporating twin bottlenecks into a unified generative network. The binary and continuous bottlenecks work collaboratively to generate discriminative binary codes without much loss of reconstruction capability.
	\vspace{-1ex}
	\item A code-driven adjacency graph is proposed with efficient computation in the Hamming space. The graph is updated adaptively to better fit the inherent data structure. Moreover, GCNs are leveraged to further exploit the data relationships.
	
	%\textbf{3)} By disentangling continuous variables from binary ones, the whole network still possesses the ability for high-quality reconstruction.
	\vspace{-1ex}
	%\item The proposed TBH model is extremely convenient for training, only requiring SGD for parameter optimization.
	\item The auto-encoding framework is novelly leveraged to determine the \textbf{reward} of the encoding quality on top of the code-driven graph, shaping the idea of \textit{learning similarities by decoding}.
	\vspace{-1ex}
	\item Extensive experiments show that the proposed TBH model massively boosts the state-of-the-art retrieval performance on four large-scale image datasets.
\end{itemize}

%-------------------------------------------------------------------------
\section{Related Work}
Learning to hash, including the supervised and unsupervised scenario~\cite{ba,lsh,dh,itq,klsh,deepbit,dgh,agh}, has been studied for years. This work is mostly related to the graph-based approaches~\cite{sh,dgh,agh,shen2018unsupervised} and deep generative models based ones~\cite{ba,sgh,hashgan,dvb,bgan,bingan}.

% TBH is related to existing unsupervised binary representation learning methods, as well as a series of generative deep learning techniques. 
\begin{figure*}[!t]
	\begin{center}
		\includegraphics[width=\textwidth]{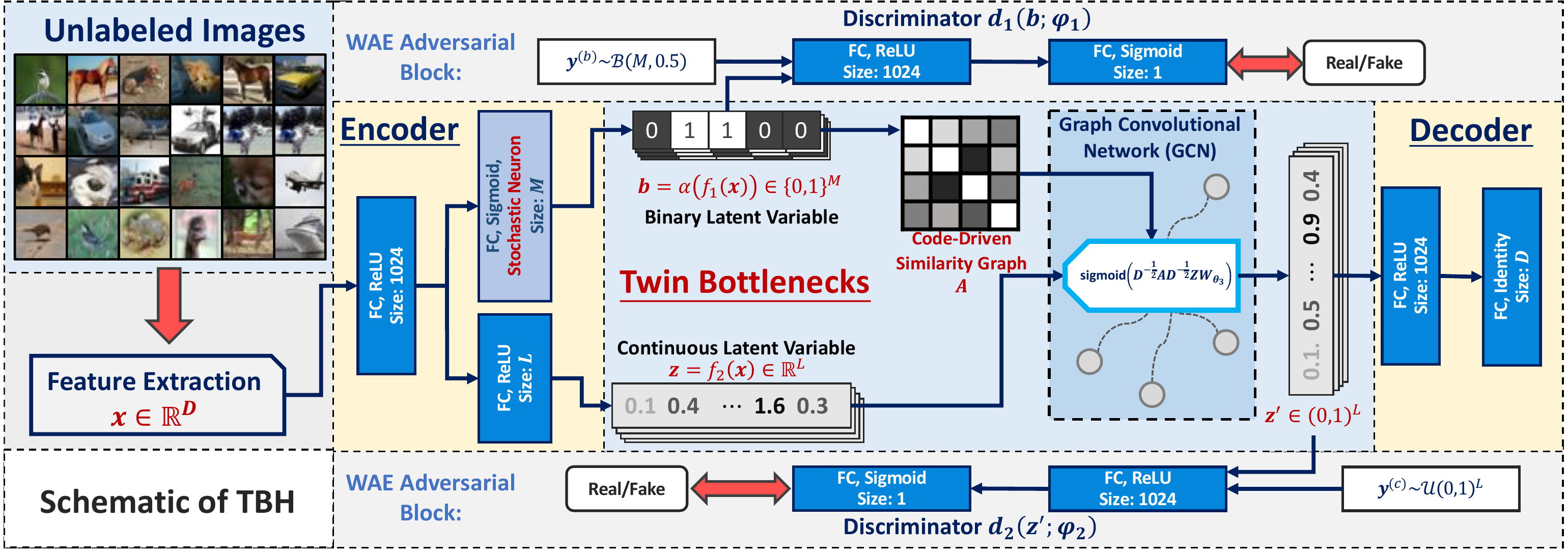}
	\end{center}
	%\vspace{-1ex}
	\caption{The schematic of TBH. % The discrete stochastic neuron produces the code $\mathbf{b}$, carrying abstracted information of data.
		Our model generally shapes an auto-encoding framework with twin bottlenecks (\ie, binary and continuous latent variables). Note that the graph adjacency $\mathbf{A}$ is dynamically built according to the code Hamming distances and is adjusted upon the \text{`reward'} from decoding. Thus, the whole process can be briefed as \textit{optimizing code similarities by decoding}.
		%After embedding the convolutional features using two kinds of encoders, we obtain both binary and continuous latent variables, \ie, $\mathbf{b}$ and $\mathbf{z}$. A code-driven similarity graph $\mathbf{A}$ is then computed based on the Hamming distances between the binary codes. $\mathbf{A}$ and $\mathbf{z}$ are simultaneously fed into the Graph Convolutional Networks (GCNs) to fully exploit the intrinsic data relevance. After decoding the similarity-preserving latent variables $\mathbf{z}'$, the updated network feedback is back-propagated to enhance the binary encoder to generate more discriminative codes. In addition, two WAE adversarial blocks are employed to implicitly regularize the twin bottlenecks.
	}
	%the top of the continuous encoder $f_2\left(\cdot\right)$. The binary encoding function $f_1\left(\cdot\right)$ is trained upon successful decoding after graph convolution. We formulate a WAE~\cite{wae} structure to implicitly regularize the latent variables. By employing distributional derivates~\cite{bengio2013estimating,sgh} on the stochastic neuron, TBH is fully solved by SGD. For testing, only $f_1\left(\cdot\right)$ is required.}
	%\vspace{-1ex}
	\label{fig_2}
\end{figure*}
\noindent\textbf{Unsupervised hashing with graphs.}~As a well-known graph-based approach, Spectral Hashing (SpH)~\cite{sh} determines pair-wise code distances according to the graph Laplacians of the data similarity affinity in the original feature space. Anchor Graph Hashing (AGH)~\cite{agh} successfully defines a small set of anchors to approximate this graph. These approaches assume that the original or mid-level data feature distance reflects the actual data relevance. As discussed in the problem of `\textbf{static graph}', this is not always realistic. Additionally, the pre-computed graph is isolated from the training process, making it hard to obtain optimal codes. Although this issue has already been considered in \cite{shen2018unsupervised} by an alternating code updating scheme, its similarity graph is still only dependent on real-valued features during training. We decide to build the batch-wise graph directly upon the Hamming distance so that the learned graph is automatically optimized by the neural network.

\noindent\textbf{Unsupervised generative hashing.}~Stochastic Generative Hashing (SGH)~\cite{sgh} is closely related to our model in the way of employing the auto-encoding framework and the discrete stochastic neurons. SGH~\cite{sgh} simply utilizes the binary latent variables as the encoder-decoder bottleneck. This design does not fully consider the code similarity and may lead to high reconstruction error, which harms the training effectiveness (`\textbf{deficient BinBN}'). Though auto-encoding schemes apply deterministic decoding error, we are also aware that some existing models~\cite{hashgan,bgan,bingan} are proposed with implicit reconstruction likelihood such as the discriminator in Generative Adversarial Network (GAN)~\cite{gan}. Note that TBH also involves adversarial training, but only for regularization purpose. %However, since TBH follows the WAE~\cite{wae} framework, the adversarial steps are only used for latent variable regularization, instead of determining reconstruction quality in \cite{hashgan,bgan,bingan}.

\section{Proposed Model}
TBH produces binary features of the given data for efficient ANN search. 
Given a data collection $\mathbf{X}=\{\mathbf{x}_i\}_{i=1}^N\in\mathbb{R}^{N\times D}$, the goal is to learn an encoding function $h:\mathbb{R}^D\rightarrow(0,1)^M$. Here $N$ refers to the set size; $D$ indicates the original data dimensionality and $M$ is the target code length. Traditionally, the code of a data point, \eg, an image or a feature vector, is obtained by applying an element-wise sign function (\ie, $\operatorname{sign}\left(\cdot\right)$) to the encoding function:
\begin{equation}\label{eq_1}
	\mathbf{b} = \left(\operatorname{sign}(h(\mathbf{x})-0.5)+1\right)/2\in\{0, 1\}^M,
\end{equation}
where $\mathbf{b}$ is the binary code. %Note that Eq.~(\ref{eq_1}) is an indifferentiable operation in the context of a neural network, and not all existing models perform quantization according to it during training. 
Some auto-encoding hashing methods~\cite{sgh,zsih} introduce stochasticity on the encoding layer (see Eq.~(\ref{eq_2})) to estimate the gradient across $\mathbf{b}$. We also adopt this paradigm to make TBH fully trainable with SGD, while during test, Eq.~(\ref{eq_1}) is used to encode out-of-sample data points. 

%The rest of this section is organized as follows. We firstly describe the network structure with each component in detail. Then the training procedure together with the learning objective and gradient estimation is introduced. Finally, the out-of-sample encoding extension of TBH is discussed.
\subsection{Network Overview}

The network structure of TBH is illustrated in Fig.~\ref{fig_2}. It typically involves a twin-bottleneck auto-encoder for our unsupervised feature learning task and two WAE~\cite{wae} discriminators for latent variable regularization. The network setting, including the numbers of layers and hidden states, is also provided in Fig.~\ref{fig_2}.

An arbitrary datum $\mathbf{x}$ is firstly fed to the encoders to produce two sets of latent variables, \ie, the binary code $\mathbf{b}$ and the continuous feature $\mathbf{z}$. Note that the back-propagatable discrete stochastic neurons are introduced to obtain the binary code $\mathbf{b}$. This procedure will be explained in Sec.~\ref{sec_32}. 
Subsequently, a similarity graph within a training batch is built according to the Hamming distances between binary codes. As shown in Fig.~\ref{fig_2}, we use an adjacency matrix $\mathbf{A}$ to represent this batch-wise similarity graph. The continuous latent variable $\mathbf{z}$ is tweaked by Graph Convolutional Network (GCN)~\cite{gcn} with the adjacency $\mathbf{A}$, resulting in the final latent variable $\mathbf{z}'$ (see Sec.~\ref{sec_33}) 
for reconstruction. Following \cite{wae}, two discriminators $d_1\left(\cdot\right)$ and $d_2\left(\cdot\right)$ are utilized to regularize the latent variables, producing informative 0-1 balanced codes.

\subsubsection{Why Does This Work?}
Our key idea is to utilize the reconstruction loss of the auxiliary decoder side as sort of \textbf{reward/critic} to score the encode quality through the GCN layer and encoder. 
Hence, TBH directly solves both `\textbf{static graph}' and `\textbf{deficient BinBN}' problems. First of all, the utilization of continuous latent variables mitigates the information loss on the binary bottleneck in \cite{ba,sgh}, as more detailed data information can be kept. This design promotes the reconstruction quality and training effectiveness. Secondly, a direct back-propagation pathway from the decoder to the binary encoder $f_1\left(\cdot\right)$ is established through the GCN~\cite{gcn}. The GCN layer selectively mixes and tunes the latent data representation based on code distances so that data with similar binary representations have stronger influences on each other. Therefore, the binary encoder is effectively trained upon successfully detecting relevant data for reconstruction.

\subsection{Auto-Encoding Twin-Bottleneck}\label{sec_320}
\subsubsection{Encoder: Learning Factorized Representations}\label{sec_32}\vspace{-1ex}
Different from conventional auto-encoders, TBH involves a \textit{twin-bottleneck} architecture. Apart from the $M$-bit binary code $\mathbf{b}\in\{0,1\}^M$, the continuous latent variable $\mathbf{z}\in\mathbb{R}^L$ is introduced to capture detailed data information. Here $L$ refers to the dimensionality of $\mathbf{z}$. As shown in Fig.~\ref{fig_2}, two encoding layers, respectively for $\mathbf{b}$ and $\mathbf{z}$, are topped on the identical fully-connected layer which receives original data representations $\mathbf{x}$. We denote these two encoding functions, \ie, $[\mathbf{b}, \mathbf{z}]=f(\mathbf{x}; \theta)$, as follows:
\vspace{-0ex}\begin{equation}\label{eq_2}\vspace{-0ex}
	\begin{split}
		\mathbf{b}&=\alpha\left(f_1(\mathbf{x}; \theta_1),\mathbf{\epsilon}\right)\in\{0,1\}^M,\\
		\mathbf{z}&=f_2\left(\mathbf{x}; \theta_2\right)\in\mathbb{R}^L,
	\end{split}
\end{equation}
where $\theta_1$ and $\theta_2$ indicate the network parameters. Note that $\theta_1$ overlaps with $\theta_2$ \wrt~the weights of the shared fully-connected layer. The first layer of $f_1\left(\cdot\right)$ and $f_2\left(\cdot\right)$ comes with a ReLU~\cite{relu} non-linearity. The activation function for the second layer of $f_1\left(\cdot\right)$ is the sigmoid function to restrict its output values within an interval of $(0,1)$, while $f_2\left(\cdot\right)$ uses ReLU~\cite{relu} non-linearity again on the second layer.

More importantly, $\alpha\left(\cdot,\mathbf{\epsilon}\right)$ in Eq.~(\ref{eq_2}) is the element-wise discrete stochastic neuron activation~\cite{sgh} with a set of random variables $\epsilon\sim\mathcal{U}\left(0,1\right)^M$, which is used for back-propagation through the binary variable $\mathbf{b}$. A discrete stochastic neuron is defined as:
\begin{equation}\label{eq_3}
	%\begin{split}
	\mathbf{b}^{(i)}=\alpha\left(f_1(\mathbf{x}; \theta_1),\epsilon\right)^{(i)}=
	\begin{cases}
		1&{f_1(\mathbf{x}; \theta_1)^{(i)}\geqslant\epsilon^{(i)},}\\
		0&{f_1(\mathbf{x}; \theta_1)^{(i)}<\epsilon^{(i)},}
	\end{cases}%\\ \text{for~} i&=1...M.
	%\end{split}
\end{equation}
where the superscript $(i)$ denotes the $i$-th element in the corresponding vector. During the training phase, this operation preserves the binary constraints and allows gradient estimation through distributional  derivative~\cite{sgh} with Monte Carlo sampling, which will be elaborated later.

\vspace{-1ex}\subsubsection{Bottlenecks: Code-Driven Hamming Graph}\label{sec_33}
%\vspace{-1ex}
Different from the existing graph-based hashing approaches \cite{dgh,agh,gcnh} where graphs are basically \textit{fixed} during training, TBH automatically detects relevant data points in a graph and mixes their representations for decoding with a back-propagatable scheme.

The outputs of the encoder, \ie, $\mathbf{b}$ and $\mathbf{z}$, are utilized to produce the final input $\mathbf{z}'$ to the decoder. For simplicity, we use batch-wise notations with capitalized letters. In particular, $\mathbf{Z}_B=[\mathbf{z}_1;\mathbf{z}_2;\cdots;\mathbf{z}_{N_B}]\in\mathbb{R}^{N_B\times L}$ and $\mathbf{B}_B=[\mathbf{b}_1;\mathbf{b}_2;\cdots;\mathbf{b}_{N_B}]\in\{0,1\}^{N_B\times M}$ respectively refer to the continuous and binary latent variables for a batch of $N_B$ data points. The inputs to the decoder are therefore denoted as $\mathbf{Z}'_B=[\mathbf{z}'_1;\mathbf{z}'_2; \cdots ;\mathbf{z}'_{N_B}]\in\mathbb{R}^{N_B\times L}$. 
We construct the graph based on the whole training batch with each datum as a vertex, and the edges are determined by the Hamming distances between the binary codes. The normalized graph adjacency $\mathbf{A}\in[0,1]^{N_B\times N_B}$ is computed by:
\begin{equation}\label{eq_a}
	\mathbf{A}=\mathbb{J}+\frac{1}{M}\left(\mathbf{B}_B(\mathbf{B}_B-\mathbb{J})^\top+(\mathbf{B}_B-\mathbb{J})\mathbf{B}_B^\top\right),
\end{equation}
where $\mathbb{J}=\mathbbm{1}^\top\mathbbm{1}$ is a matrix full of ones. Eq.~(\ref{eq_a}) is an equilibrium of $\mathbf{A}_{ij} = 1-\operatorname{Hamming}(\mathbf{b}_i, \mathbf{b}_j)/M$ for each entry of $\mathbf{A}$. Then this adjacency, together with the continuous variables $\mathbf{Z}_B$, is processed by the GCN layer~\cite{gcn}, which is defined as:
\vspace{-0ex}\begin{equation}\label{eq_gcn}\vspace{-0ex}
	\mathbf{Z}'_B= \operatorname{sigmoid}\big(\mathbf{D}^{-\frac{1}{2}}\mathbf{A}\mathbf{D}^{-\frac{1}{2}}\mathbf{Z}_B\mathbf{W}_{\theta_3}\big).
\end{equation}
Here $\mathbf{W}_{\theta_3}\in\mathbb{R}^{L\times L}$ is a set of trainable projection parameters and $\mathbf{D}=\operatorname{diag}\left(\mathbf{A}\mathbbm{1}^\top\right)$. %As discussed in \cite{gcn}, Eq.~(\ref{eq_gcn}) is the first-order approximation of the Chebyshev polynomials of a parameterized convolution operation $k_{\theta_3}\left(\mathbf{B}_B\right)\ast\mathbf{Z}_B$ and it works well as a part of a neural network.

As the batch-wise adjacency $\mathbf{A}$ is constructed exactly from the codes, a trainable pathway is then established from the decoder to $\mathbf{B}_B$. Intuitively, the reconstruction penalty scales up when unrelated data are closely located in the Hamming space. Ultimately, only relevant data points with similar binary representations are linked during decoding. Although GCNs~\cite{gcn} are utilized as well in \cite{zsih,gcnh}, these works generally use pre-computed graphs and cannot handle the `\textbf{static graph}' problem.

\vspace{-1ex}\subsubsection{Decoder: Rewarding the Hashing Results}\label{sec_32b}
The decoder is an auxiliary component to the model, determining the code quality produced by the encoder. As shown in Fig.~\ref{fig_2}, the decoder $g\left(\cdot\right)$ of TBH consists of two fully-connected layers, which are topped on the GCN~\cite{gcn} layer. We impose an ReLU~\cite{relu} non-linearity on the first layer and an identity activation on the second layer. Therefore, the decoding output $\hat{\mathbf{x}}$ is represented as $\hat{\mathbf{x}}=g\left(\mathbf{z}';\theta_4\right)\in\mathbb{R}^D$, where $\theta_4$ refers to the network parameters within the scope of the decoder and $\mathbf{z}'$ is a row vector of $\mathbf{Z}'_B$ generated by the GCN~\cite{gcn}. We elaborate the detailed loss of the decoding side in Sec.~\ref{sec_35}.

%As a conventional solution to unsupervised learning, auto-encoding networks have also been adopted in \cite{ba,sgh}. 
To keep the content concise, we do not propose a large convolutional network receiving and generating raw images, since our goal is to learn compact binary features. The decoder provides deterministic reconstruction penalty, \eg, the $\ell_2$-norm, back to the encoders during optimization. This ensures a more stable and controllable training procedure than the implicit generation penalties, \eg, the discriminators in GAN-based hashing~\cite{hashgan,bgan,bingan}. %The input data representations of TBH can be obtained by any arbitrary models.

\subsection{Implicit Bottleneck Regularization}\label{sec_34}
The latent variables in the bottleneck are regularized to avoid wasting bits and align representation distributions. Different from the deterministic regularization terms such as bit de-correlation~\cite{dh,deepbit} and entropy-like loss~\cite{sgh}, TBH mimics WAE~\cite{wae} to adversarially regularize the latent variables with auxiliary discriminators. 
The detailed settings of the discriminators, \ie, $d_1\left(\cdot;\varphi_1\right)$ and $d_2\left(\cdot;\varphi_2\right)$ with network parameters $\varphi_1$ and $\varphi_2$, are illustrated in Fig.~\ref{fig_2}, particularly involving two fully-connected layers successively with ReLU~\cite{relu} and sigmoid non-linearities.

In order to balance zeros and ones in a binary code, we assume that $\mathbf{b}$ is priored by a binomial distribution $\mathcal{B}\left(M,0.5\right)$, which could maximize the code entropy. Meanwhile, regularization is also applied to the continuous variables $\mathbf{z}'$ after the GCN for decoding. We expect $\mathbf{z}'$ to obey a uniform distribution $\mathcal{U}\left(0,1\right)^L$ to fully explore the latent space. To that end, we employ the following two discriminators $d_1$ and $d_2$ for $\mathbf{b}$ and $\mathbf{z}'$, respectively:
\begin{equation}
	\begin{split}
		d_1\left(\mathbf{b};\varphi_1\right)\in(0,1)&;\ d_1(\mathbf{y}^{(b)};\varphi_1)\in(0,1),\\
		d_2\left(\mathbf{z}';\varphi_2\right)\in(0,1)&;\ d_2(\mathbf{y}^{(c)};\varphi_2)\in(0,1),
	\end{split}
\end{equation}
where $\mathbf{y}^{(b)}\sim\mathcal{B}\left(M,0.5\right)$ and $\mathbf{y}^{(c)}\sim\mathcal{U}\left(0,1\right)^L$ are random signals sampled from the targeted distributions for implicit regularizing $\mathbf{b}$ and $\mathbf{z}'$ respectively. %The adversarial learning objectives are derived in Sec~\ref{sec_35}.

The WAE-like regularizers focus on minimizing the distributional discrepancy between the produced feature space and the target one. This design fits TBH better than deterministic regularizers~\cite{sgh,dh}, since such kinds of regularizers (\eg, bit de-correlation) impose direct penalties on each sample, which may heavily skew the similarity graph built upon the codes and consequently degrades the training quality. Experiments also support our insight (see Table \ref{tab_abl}).

%\vspace{-1ex}
\subsection{Learning Codes and Similarity by Decoding}\label{sec_35}
As TBH involves adversarial training, two learning objectives, \ie, $\mathcal{L}_{AE}$ for the \textit{auto-encoding} step and $\mathcal{L}_D$ for the \textit{discriminating} step, are respectively introduced.

\vspace{-2ex}\subsubsection{\textit{Auto-Encoding} Objective}
The \textit{Auto-Encoding} objective $\mathcal{L}_{AE}$ is written as follows:
\vspace{-1ex}\begin{equation}\label{eq_lae}\vspace{-0ex}
	\begin{split}\vspace{-1ex}
		\mathcal{L}_{AE}=\frac{1}{N_B}&\sum_{i=1}^{N_B}\mathbb{E}_\mathbf{b_i}\Big[\frac{1}{2M}\|\mathbf{x}_i-\hat{\mathbf{x}}_i\|^2\\
		&-\lambda\log d_1(\mathbf{b}_i;\varphi_1)-\lambda\log d_2(\mathbf{z}'_i;\varphi_2)\Big],
	\end{split}
\end{equation}
where $\lambda$ is a hyper-parameter controlling the penalty of the discriminator according to \cite{wae}. $\mathbf{b}$ is obtained from Eq.~(\ref{eq_3}), $\mathbf{z}'$ is computed according to Eq.~(\ref{eq_gcn}), and the decoding result $\hat{\mathbf{x}}=g\left(\mathbf{z}';\theta_4\right)$ is obtained from the decoder. $\mathcal{L}_{AE}$ is used to optimize the network parameters within the scope of $\theta=\{\theta_1,\theta_2,\theta_3,\theta_4\}$. Eq.~(\ref{eq_lae}) comes with an expectation term $\mathbb{E}[\cdot]$ over the latent binary code, since $\mathbf{b}$ is generated by a sampling process.

Inspired by~\cite{sgh}, we estimate the gradient through the binary bottleneck with distributional derivatives by utilizing a set of random signals $\epsilon\sim\mathcal{U}\left(0,1\right)^M$. The gradient of $\mathcal{L}_{AE}$ \wrt~$\theta$ is estimated by:
\vspace{-1ex}\begin{equation}\label{eq_gae}\vspace{-0ex}
	\begin{split}
		\nabla_\theta\mathcal{L}_{AE}\approx&\frac{1}{N_B}\sum_{i=1}^{N_B}\mathbb{E}_\epsilon\Big[\nabla_\theta\big(\frac{1}{2M}\|\mathbf{x}_i-\hat{\mathbf{x}}_i\|^2\\
		&-\lambda\log d_1(\mathbf{b}_i;\varphi_1)-\lambda\log d_2(\mathbf{z}'_i;\varphi_2)\big)\Big].
	\end{split}
\end{equation}
We refer the reader to \cite{sgh} and our \textbf{Supplementary Material} for the details of Eq.~\eqref{eq_gae}. Notably, a similar approach of sampling-based gradient estimation for discrete variables was employed in \cite{bengio2013estimating}, which has been proved as a special case of the REINFORCE algorithm~\cite{reinforce}.

\vspace{-2ex}\begin{algorithm}[t]
	\small
	\caption{The Training Procedure of TBH}
	\label{alg}
	\textbf{Input:}\hspace{0mm} Dataset $\mathbf{X}=\{\mathbf{x}_i\}_{i=1}^N$ and the maxinum number of iterations $T$.\\
	\textbf{Output:}\hspace{0mm} Network parameters $\theta_1$.\\
	%	\BlankLine
	%Randomly initialize $\mathbf{H}\in\{-1,1\}^{M\times N}$\\
	\Repeat{convergence or reaching the maximum iteration $T$}{
		Randomly select a mini-batch $\{\mathbf{x}_i\}_{i=1}^{N_B}$ from $\mathbf{X}$\\
		\For{$i=1, \cdots, N_B$}{
			Sample $\epsilon_i\sim\mathcal{U}\left(0,1\right)^M$\\
			Sample $\mathbf{y}_i^{(b)}\sim\mathcal{B}\left(M,0.5\right);\mathbf{y}_i^{(c)}\sim\mathcal{U}\left(0,1\right)^L$\\
			Sample a set of $\mathbf{b}_i$ according to Eq.~(\ref{eq_3}) using $\epsilon_i$}
		\textbf{\textit{Discriminating} step:}\\
		\hspace{4.3mm} $\mathcal{L}_{D}\leftarrow$ Eq.~(\ref{eq_ld})\\
		\hspace{4.3mm} $\varphi\leftarrow\varphi-\mathbf{\Gamma}\left(\nabla_\varphi\mathcal{L}_{D}\right)$\\
		\textbf{\textit{Auto-encoding} step:}\\
		\hspace{4.3mm} $\mathcal{L}_{AE}\leftarrow$ Eq.~(\ref{eq_lae})\\
		\hspace{4.3mm} $\theta\leftarrow\theta-\mathbf{\Gamma}\left(\nabla_\theta\mathcal{L}_{AE}\right)$ according to Eq.~(\ref{eq_gae})
	}
\end{algorithm}
\subsubsection{\textit{Discriminating} Objective}\vspace{-1ex}
The \textit{Discriminating} objective $\mathcal{L}_D$ is defined by:
\vspace{-1ex}
\begin{equation}\label{eq_ld}
	\begin{split}
		\mathcal{L}_D=&-\frac{\lambda}{N_B}\sum_{i=1}^{N_B}\Big(\log d_1(\mathbf{y}_i^{(b)}; \varphi_1) \\
		&+\log d_2(\mathbf{y}_i^{(c)}; \varphi_2)+\log\big(1-d_1(\mathbf{b}_i;\varphi_1)\big)\\ 
		&+\log\big(1-d_2(\mathbf{z}'_i;\varphi_2)\big)
		\Big).
	\end{split}
\end{equation}
\vspace{0ex}Here $\lambda$ refers to the same hyper-parameter as in Eq.~(\ref{eq_lae}). Similarly, $\mathcal{L}_{D}$ optimizes the network parameters within the scope of $\varphi=\{\varphi_1,\varphi_2\}$. As the \textit{discriminating} step does not propagate error back to the auto-encoder, there is no need to estimate the gradient through the indifferentiable binary bottleneck. Thus the expectation term $\mathbb{E}[\cdot]$ in Eq.~(\ref{eq_lae}) is deprecated in Eq.~(\ref{eq_ld}).

The training procedure of TBH is summarized in Algorithm~\ref{alg}, where $\mathbf{\Gamma}\left(\cdot\right)$ refers to the adaptive gradient scaler, for which we adopt the Adam optimizer~\cite{adam}. Monte Carlo sampling is performed on the binary bottleneck, once a data batch is fed to the encoder. Therefore, the learning objective can be computed using the network outputs. %All parameters are optimized with the standard SGD algorithm.

\subsection{Out-of-Sample Extension}\label{sec_36}
After TBH is trained, we can obtain the binary codes for any out-of-sample data as follows:
\begin{equation}\label{eq_ose}
	\mathbf{b}^{(q)}=(\operatorname{sign}(f_1(\mathbf{x}^{(q)};\theta_1)-0.5)+1)/2\in\{0,1\}^{M},
\end{equation}
where $\mathbf{x}^{(q)}$ denotes a query data point. During the test phase, only $f_1\left(\cdot\right)$ is required, which considerably eases the binary coding process. Since only the forward propagation process is involved for test data, the stochasticity on the encoder $f_1\left(\cdot\right)$ used for training in Eq. \eqref{eq_2} is not needed.

\begin{table*}[t]
	\begin{center}
		\caption{Performance comparison (\wrt~MAP) of TBH and the state-of-the-art \textbf{unsupervised} hashing methods. %Note that TBH is \textcolor{blue!80!black}{\textbf{unsupervised}}. The retrieval performance of some supervised hashing methods is reported only for better illustration.
		}\label{tab_map}\vspace{-1ex}
		\small
		\resizebox{\textwidth}{!}{
			\begin{tabular}{l l ccc  ccc  ccc}
				\hline
				\rowcolor{gray!15}&&\multicolumn{3}{c}{\textbf{ CIFAR-10}}&\multicolumn{3}{c}{\textbf{NUS-WIDE}}&\multicolumn{3}{c}{\textbf{MSCOCO}}\\\hline
				\rowcolor{gray!15}\textbf{Method}&\textbf{Reference}& 16 bits& 32 bits & 64 bits& 16 bits& 32 bits & 64 bits& 16 bits& 32 bits & 64 bits\\ \hline\hline
				LSH~\cite{lsh}&STOC02& 0.106& 0.102 & 0.105 & 0.239& 0.266 & 0.266& 0.353& 0.372& 0.341\\%& 0.152 & 0.141 & 0.165\\
				SpH~\cite{sh}&NIPS09& 0.272& 0.285& 0.300 & 0.517& 0.511& 0.510 & 0.527& 0.529& 0.546\\%& 0.185& 0.271& 0.350\\
				AGH~\cite{agh}&ICML11& 0.333& 0.357 & 0.358 & 0.592& 0.615& 0.616&0.596 &0.625 & 0.631\\%& 0.241& 0.326& 0.379\\
				SpherH~\cite{sph}&CVPR12& 0.254 & 0.291 & 0.333 & 0.495 & 0.558 & 0.582 & 0.516& 0.547 &0.589\\%& 0.110 & 0.187& 0.259\\
				% PCAH&\xmark& 21.52& 21.62& 20.54& & & & & & & & & \\
				KMH~\cite{he2013k}&CVPR13& 0.279 & 0.296& 0.334 & 0.562& 0.597 & 0.600 & 0.543 & 0.554 & 0.592\\% & 0.202 & 0.297 & 0.390\\
				ITQ~\cite{itq}&PAMI13& 0.305& 0.325& 0.349& 0.627& 0.645& 0.664& 0.598& 0.624& 0.648\\%& 0.217& 0.317& 0.391\\
				DGH~\cite{dgh}&NIPS14& 0.335& 0.353& 0.361& 0.572& 0.607& 0.627& 0.613& 0.631&0.638\\\hline% &0.270 &0.348 &0.373\\\cline{1-13}
				% DH&\xmark& 16.17& 16.62& 16.96& & 48& & & & & & &\\
				DeepBit~\cite{deepbit}&CVPR16& 0.194& 0.249& 0.277& 0.392& 0.403& 0.429& 0.407& 0.419 & 0.430 \\%& 0.204& 0.283 &0.286\\
				SGH~\cite{sgh}&ICML17 & 0.435& 0.437 & 0.433& 0.593& 0.590& 0.607& 0.594& 0.610& 0.618\\%& 0.447& 0.500& 0.523\\
				BGAN~\cite{bgan}&AAAI18& 0.525& 0.531& 0.562& 0.684& 0.714& 0.730& 0.645& 0.682& 0.707\\%& 0.499& 0.535& 0.574\\
				BinGAN~\cite{bingan}&NIPS18& 0.476& 0.512& 0.520& 0.654& 0.709& 0.713& 0.651& 0.673& 0.696\\%& 0.505& 0.584& 0.612\\
				GreedyHash~\cite{greedyhash}&NIPS18& 0.448& 0.473& 0.501& 0.633& 0.691& 0.731 & 0.582& 0.668& 0.710\\%& 0.186& 0.578& 0.558\\
				HashGAN~\cite{hashgan}*&CVPR18& 0.447& 0.463& 0.481& -& -& -& -& -& -\\%& -& -& -\\
				DVB~\cite{dvbj}&IJCV19& 0.403& 0.422& 0.446& 0.604& 0.632& 0.665& 0.570& 0.629& 0.623\\%& 0.398& 0.452& 0.465\\
				DistillHash~\cite{distill}& CVPR19& 0.284& 0.285& 0.288& 0.667& 0.675& 0.677& -& -& -\\%\hline\hline%& -& -& -\\\hline\hline
				\hline
				\rowcolor{gray!15}\textbf{TBH}& Proposed&\textbf{0.532}&\textbf{0.573}&\textbf{0.578}&\textbf{0.717}&\textbf{0.725}&\textbf{0.735}&\textbf{0.706}&\textbf{0.735}&\textbf{0.722}\\\hline%&\textbf{0.560}&\textbf{0.619}&\textbf{0.626}\\\hline
			\end{tabular}
		}
		\\\small{*Note the duplicate naming of HashGAN, \ie, the unsupervised one~\cite{hashgan} and the supervised one~\cite{shashgan}.}
	\end{center}\vspace{-4ex}
\end{table*}
\vspace{-1ex}\section{Experiments}\vspace{-1ex}

We evaluate the performance of the proposed TBH on four large-scale image benchmarks, \ie, CIFAR-10, NUS-WIDE, MS COCO. %, and ImageNet. 
We additionally present results for image reconstruction on the MNIST dataset. 
% and person re-identification results on the Market-1501 dataset, showing the further potential applications of TBH.
\subsection{Implementation Details}
% 	\begin{figure}[t]
% 	\begin{center}
% 		\includegraphics[width=\linewidth]{eff_cmp2.pdf}
% 	\end{center}\vspace{-2ex}
% 	\caption{Retrieval MAP@1000 results with extremely short code length $M=4, \cdots, 12$ on CIFAR-10~\cite{cifar}.}%\vspace{-1ex}
% 	\label{fig_short}
% 	\end{figure}

The proposed TBH model is implemented with the popular deep learning toolbox Tensorflow~\cite{tf}. The hidden layer sizes and the activation functions used in TBH are all provided in Fig.~\ref{fig_2}. The gradient estimation of Eq.~(\ref{eq_gae}) can be implemented with a single Tensorflow decorator in Python, following \cite{sgh}. TBH only involves \textbf{two} hyper-parameters, \ie, $\lambda$ and $L$. We set $\lambda=1$ and $L=512$ by default. For all of our experiments, the \texttt{fc\_7} features of the AlexNet~\cite{alexnet} network are utilized for data representation. The learning rate of Adam optimizer $\mathbf{\Gamma}\left(\cdot\right)$~\cite{adam} is set to $1\times10^{-4}$, with default decay rates $\mathtt{beta}_1=0.9$ and $\mathtt{beta}_2=0.999$. We fix the training batch size to 400. Our implementation can be found at \url{https://github.com/ymcidence/TBH}.% and report the results at the 20-th epoch.

\subsection{Datasets and Setup}
\begin{figure*}
	\begin{center}
		\includegraphics[width=0.99\textwidth]{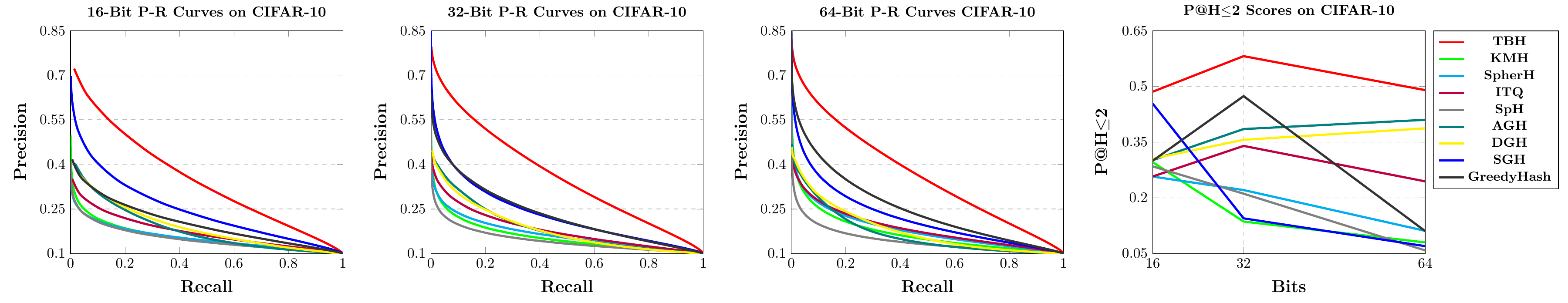}        
	\end{center}\vspace{-2ex}
	%\vspace{-1ex}
	\caption{P-R curves and P@H$\leq$2 results of TBH and compared methods on CIFAR-10.}\vspace{-2ex}
	\label{fig_pr}
\end{figure*}

\noindent\textbf{CIFAR-10~\cite{cifar}} consists of 60,000 images from 10 classes. We follow the common setting \cite{hashgan,greedyhash} and select 10,000 images (1000 per class) as the query set. The remaining 50,000 images are regarded as the database.%, 5,000 (500 per class) of which are employed for training.

\noindent\textbf{NUS-WIDE~\cite{nus}} is a collection of nearly 270,000 images of 81 categories. Following the settings in \cite{cnnh}, we adopt the subset of images from 21 most frequent categories. 100 images of each class are utilized as a query set and the remaining images form the database. For training, we employ 10,500 images uniformly selected from the 21 classes.

\noindent\textbf{MS COCO~\cite{coco}} is a benchmark for multiple tasks. We adopt the pruned set as with \cite{cao2017hashnet} with 12,2218 images from 80 categories. We randomly select 5,000 images as queries with the remaining ones the database, from which 10,000 images are chosen for training.

%\noindent\textbf{ImageNet~\cite{imagenet}} consists of 1,000 image classes of images. 
%in Large Scale Visual Recognition Challenge (ILSVRC). 
%The original training and validation sets include over 1.2M images and 50K images, respectively. 
%Following \cite{cao2017hashnet}, we select 100 categories to perform our retrieval task. All the original training images are used as the database, and all the validation images form the query set. We randomly select 130 images per category from the database for training.

Standard metrics \cite{cao2017hashnet,greedyhash} are adopted to evaluate our method and other state-of-the-art methods, \ie, Mean Average Precision (MAP), Precision-Recall (P-R) curves, Precision curves within Hamming radius 2 (P@H$\leq$2), and Precision \wrt 1,000 top returned samples (P@1000). We adopt MAP@1000 for CIFAR-10, and MAP@5000 for MS COCO and NUS-WIDE according to \cite{cao2017hashnet,dhn}.
\begin{table}[t]
	\caption{P@1000 results of TBH and compared methods on CIFAR-10 and MSCOCO.}
	\label{tab_pre}
	\small
	\resizebox{0.99\linewidth}{!}{
		\begin{tabular}{l ccc ccc}
			\hline
			\rowcolor{gray!15}&\multicolumn{3}{c}{\textbf{ CIFAR-10}}&\multicolumn{3}{c}{\textbf{MSCOCO}}\\\hline
			\rowcolor{gray!15}\textbf{Method}&16 bits& 32 bits & 64 bits& 16 bits& 32 bits & 64 bits\\\hline\hline
			KMH & 0.242& 0.252& 0.284& 0.557 & 0.572& 0.612 \\
			SpherH & 0.228& 0.256& 0.291& 0.525 & 0.571 & 0.612\\
			%LSH~\cite{lsh} & 9.95& 9.65& 10.02&0.99&0.92&1.09\\
			ITQ & 0.276& 0.292 & 0.309 & 0.607& 0.637& 0.662\\
			SpH & 0.238 & 0.239& 0.245& 0.541& 0.548& 0.567\\
			AGH & 0.306& 0.321& 0.317& 0.602& 0.635& 0.644\\
			DGH & 0.315& 0.323& 0.324 & 0.623& 0.642& 0.650\\
			HashGAN & 0.418& 0.436& 0.455 &-&-&-\\
			SGH & 0.387 & 0.380 & 0.367 & 0.604 & 0.615 & 0.637\\
			GreedyHash & 0.322 & 0.403 & 0.444 & 0.603 & 0.624 & 0.675 \\
			\hline
			\rowcolor{gray!15}\textbf{TBH (Ours)} & \textbf{0.497}& \textbf{0.524}& \textbf{0.529}& \textbf{0.646}& \textbf{0.698}& \textbf{0.701}\\\hline
		\end{tabular}
	}\vspace{-0ex}
\end{table}

\subsection{Comparison with Existing Methods}\vspace{-1ex}
%\subsubsection{Baselines} 
We compare TBH with several state-of-the-art unsupervised hashing methods, including \textbf{LSH}~\cite{lsh}, \textbf{ITQ}~\cite{itq}, \textbf{SpH}~\cite{sh}, \textbf{SpherH}~\cite{sph}, \textbf{KMH}~\cite{he2013k}, \textbf{AGH}~\cite{agh}, \textbf{DGH}~\cite{dgh}, \textbf{DeepBit}~\cite{deepbit}, \textbf{BGAN}~\cite{bgan}, \textbf{HashGAN}~\cite{hashgan}, \textbf{SGH~\cite{sgh}}, \textbf{BinGAN}~\cite{bingan}, \textbf{GreedyHash}~\cite{greedyhash}, \textbf{DVB}~\cite{dvbj} and \textbf{DistillHash}~\cite{distill}. %, but also supervised shallow/deep ones (\textbf{KSH~\cite{ksh}}, \textbf{ITQ-CCA~\cite{ccaitq}}, \textbf{SDH~\cite{sdh}}, \textbf{CNNH~\cite{cnnh}}, \textbf{DNNH~\cite{dnnh}}, \textbf{DHN~\cite{dhn}}, and \textbf{HashNet~\cite{cao2017hashnet}}). 
For fair comparisons, all the methods are reported with identical training and test sets. Additionally, the shallow methods are evaluated with the same deep features as the ones we are using.

\vspace{-1ex}\subsubsection{Retrieval results}\vspace{-1ex}
The MAP and P@1000 results of TBH and other methods are respectively provided in Tables \ref{tab_map} and \ref{tab_pre}, while the respective P-R curves and P@H$\leq$2 results are illustrated in Fig.~\ref{fig_pr}. The performance gap between TBH and existing unsupervised methods can be clearly observed. Particularly, TBH obtains remarkable MAP gain with 16-bit codes (\ie, $M=16$). Among the unsupervised baselines, GreedyHash~\cite{greedyhash} performs closely next to TBH. It bases the produced code similarity on pair-wise feature distances. As is discussed in Sec~\ref{sec_1}, this design is straightforward but sub-optimal since the original feature space is not fully revealing data relevance. On the other hand, as a generative model, HashGAN~\cite{hashgan} significantly underperforms TBH, as the binary constraints are violated during its adversarial training. TBH differs SGH~\cite{sgh} by leveraging the twin-bottleneck scheme. Since SGH~\cite{sgh} only considers the reconstruction error and in auto-encoder, it generally does not produce convincing retrieval results.

%Supervised hashing baselines are selectively reported to further demonstrate the superiority of TBH. Conventionally, it is not reasonable to expect unsupervised hashing models to outperform all existing supervised ones. However, we show that, under the experimental settings we are following, TBH reaches the performance level of several well-known supervised methods.

\vspace{-2ex}\subsubsection{Extremely short codes}
\begin{figure}[t]
	\begin{center}
		\includegraphics[width=\linewidth]{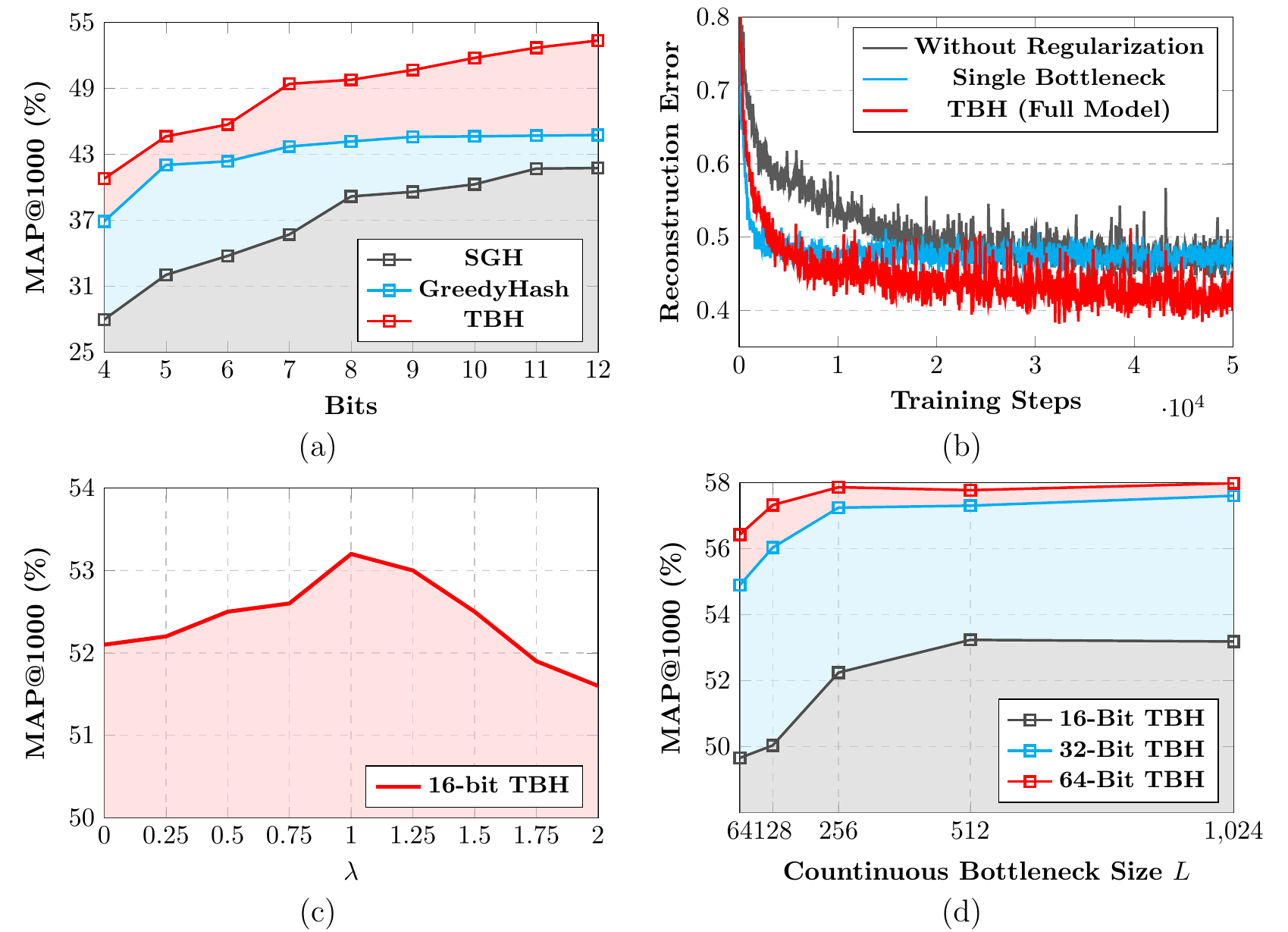}
	\end{center}
	\vspace{-2ex}
	\caption{\textbf{(a)} MAP@1000 results with extremely short code lengths on CIFAR-10. \textbf{(b)} 16-bit normalized reconstruction errors of TBH and its variants on CIFAR-10. \textbf{(c)} and \textbf{(d)} Effects of the hyper-parameters $\lambda$ and $L$ on CIFAR-10.}\vspace{-0ex}
	\label{fig_cmp}
\end{figure}

\vspace{-1ex}
Inspired by \cite{greedyhash}, we illustrate the retrieval performance with extremely short bit length in Fig.~\ref{fig_cmp} (a). TBH works well even when the code length is set to $M=4$. The significant performance gain over SGH can be observed. This is due to that, during training, the continuous bottleneck complements the information discarded by the binary one.
\begin{table}
	\caption{Ablation study of MAP@1000 results by using different variants of the proposed TBH on CIFAR-10.}
	\label{tab_abl}
	\small
	\resizebox{0.99\linewidth}{!}{
		\begin{tabular}{ll ccc}
			\hline
			\rowcolor{gray!15}&\textbf{Baseline}& \textbf{16 bits}& \textbf{32 bits}& \textbf{64 bits}\\\hline\hline
			1&Single bottleneck & 0.435& 0.437 & 0.433\\
			%2&Swapped bottlenecks& 0.470 & 0.525 & 0.517\\
			2&Swapped bottlenecks& 0.466 & 0.471 & 0.475\\
			3&Explicit regularization & 0.524& 0.559 & 0.560\\
			4&Without regularization & 0.521& 0.535& 0.547\\
			5&Without stochastic neuron& 0.408& 0.412 & 0.463\\
			6&Fixed graph & 0.442& 0.464& 0.459 \\
			7&Attention equilibrium& 0.477& 0.503& 0.519\\\hline
			\rowcolor{gray!15}&\textbf{TBH (full model)}& \textbf{0.532}& \textbf{0.573}& \textbf{0.578}\\\hline
		\end{tabular}
	}\vspace{-0ex}
\end{table}

\vspace{-1ex}\subsection{Ablation Study}\vspace{-1ex}
In this subsection, we validate the contribution of each component of TBH, and also show some empirical analysis. Different baseline network structures are visualized in the \textbf{Supplementary Material} for better understanding.

\vspace{-1ex}\subsubsection{Component Analysis}\vspace{-1ex}
%\footnote{We visualize different baseline network structures in the \textbf{Supplementary Material} for better understanding.}
We compare TBH with the following baselines. \textbf{(1) Single bottleneck.} This baseline coheres with SGH. We remove the twin-bottleneck structure and directly feed the binary codes to the decoder. 
\textbf{(2) Swapped bottlenecks.} We swap the functionalities of the two bottlenecks, \ie, using the continuous one for adjacency building and the binary one for decoding. 
\textbf{(3) Explicit regularization.} The WAE regularizers are replaced by conventional regularization terms. An entropy loss similar to SGH is used to regularize $\mathbf{b}$, while an $\ell_2$-norm is applied to $\mathbf{z}'$. 
\textbf{(4) Without regularization.} The regularization terms for $\mathbf{b}$ and $\mathbf{z}'$ are removed in this baseline. 
\textbf{(5) Without stochastic neuron.} The discrete stochastic neuron $\alpha\left(\cdot,\mathbf{\epsilon}\right)$ is deprecated on the top of $f_1(\cdot)$, and bit quantization loss~\cite{dh} is appended to $\mathcal{L}_{AE}$. 
\textbf{(6) Fixed graph.} $\mathbf{A}$ is pre-computed using feature distances. The continuous bottleneck is removed and the GCN is applied to the binary bottleneck with the fixed $\mathbf{A}$. \textbf{(7) Attention equilibrium.} This baseline performs weighted average on $\mathbf{Z}$ according to $\mathbf{A}$, instead of employing GCN in-between.

Table~\ref{tab_abl} shows the performance of the baselines. We can observe that the model undergoes significant performance drop when modifying the twin-bottleneck structure. Specifically, our trainable adaptive Hamming graph plays an important role in the network. When removing this (\ie, \texttt{baseline\_6}), the performance decreases by $\sim$9\%. This accords with our motivation in dealing with the `\textbf{static graph}' problem.In practice, we also experience training perturbations when applying different regularization and quantization penalties to the model.

\begin{figure}[!t]
	\begin{center}
		\includegraphics[width=.9\linewidth]{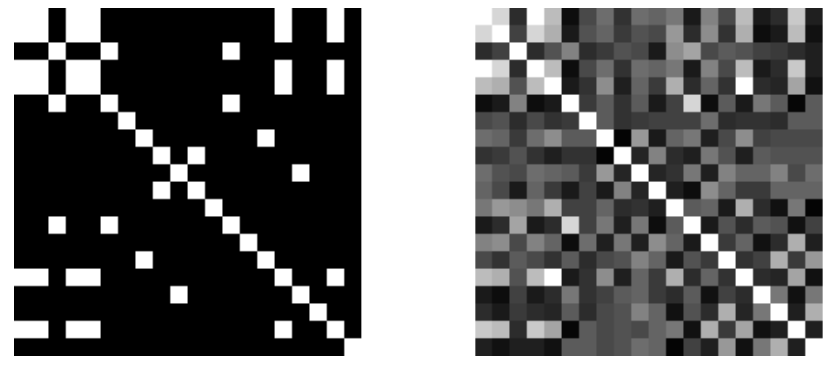}
	\end{center}\vspace{-2ex}
	%\vspace{-1ex}
	\caption{Adjacency matrices of 20 randomly-sampled data points of a training batch on CIFAR-10, computed based on ground-truth similarity (\textbf{Left}) and the Hamming distances between binary codes using Eq.~(\ref{eq_a}) (\textbf{Right}).}\vspace{-0ex}
	\label{fig_aff}
\end{figure}

\vspace{-2ex}\subsubsection{Reconstruction Error}\vspace{-1ex} 
As mentioned in the `\textbf{deficient BinBN}' problem, decoding from a single binary bottleneck is less effective. This is illustrated in Fig.~\ref{fig_cmp} (b), where the normalized reconstruction errors of TBH, \texttt{baseline\_1} and \texttt{baseline\_4} are plotted. TBH produces lower decoding error than the single bottleneck baseline. Note that \texttt{baseline\_1} structurally coheres SGH~\cite{sgh}. 
%We also report the retrieval MAP \wrt training steps of TBH in Fig.~\ref{fig_cmp} (c) to better demonstrate the effectiveness issue.

\vspace{-1ex}\subsubsection{Hyper-Parameter}\vspace{-1ex}
Only two hyper-parameters are involved in TBH. The effect of the adversarial loss scaler $\lambda$ is illustrated in Fig.~\ref{fig_cmp} (c). A large regularization penalty slightly influences the overall retrieval performance. The results \wrt~different settings of $L$ on CIFAR-10 are shown in Fig.~\ref{fig_cmp} (d). Typically, no dramatic performance drop is observed when squeezing the bottleneck, as data relevance is not only reflected by the continuous bottleneck. Even when setting $L$ to 64, TBH still outperforms most existing unsupervised methods, which also endorses our twin-bottleneck mechanism.

\subsection{Qualitative Results}\vspace{-1ex}

We provide some intuitive results to further justify the design. The implementation details are given in the \textbf{Supplementary Material} to keep the content concise.

\vspace{-1ex}\subsubsection{The Constructed Graph by Hash Codes}\vspace{-1ex}
We show the effectiveness of the code-driven graph learning process in Fig.~\ref{fig_aff}. 20 random samples are selected from a training batch to plot the adjacency. The twin-bottleneck mechanism automatically tunes the codes, constructing $\mathbf{A}$ based on Eq.~(\ref{eq_a}). Though TBH has no access to the labels, the constructed adjacency simulates the label-based one. Here brighter color indicates closer Hamming distances.

\vspace{-1ex}\subsubsection{Visualization}
\begin{figure}[t]
	\begin{center}
		\includegraphics[width=\linewidth]{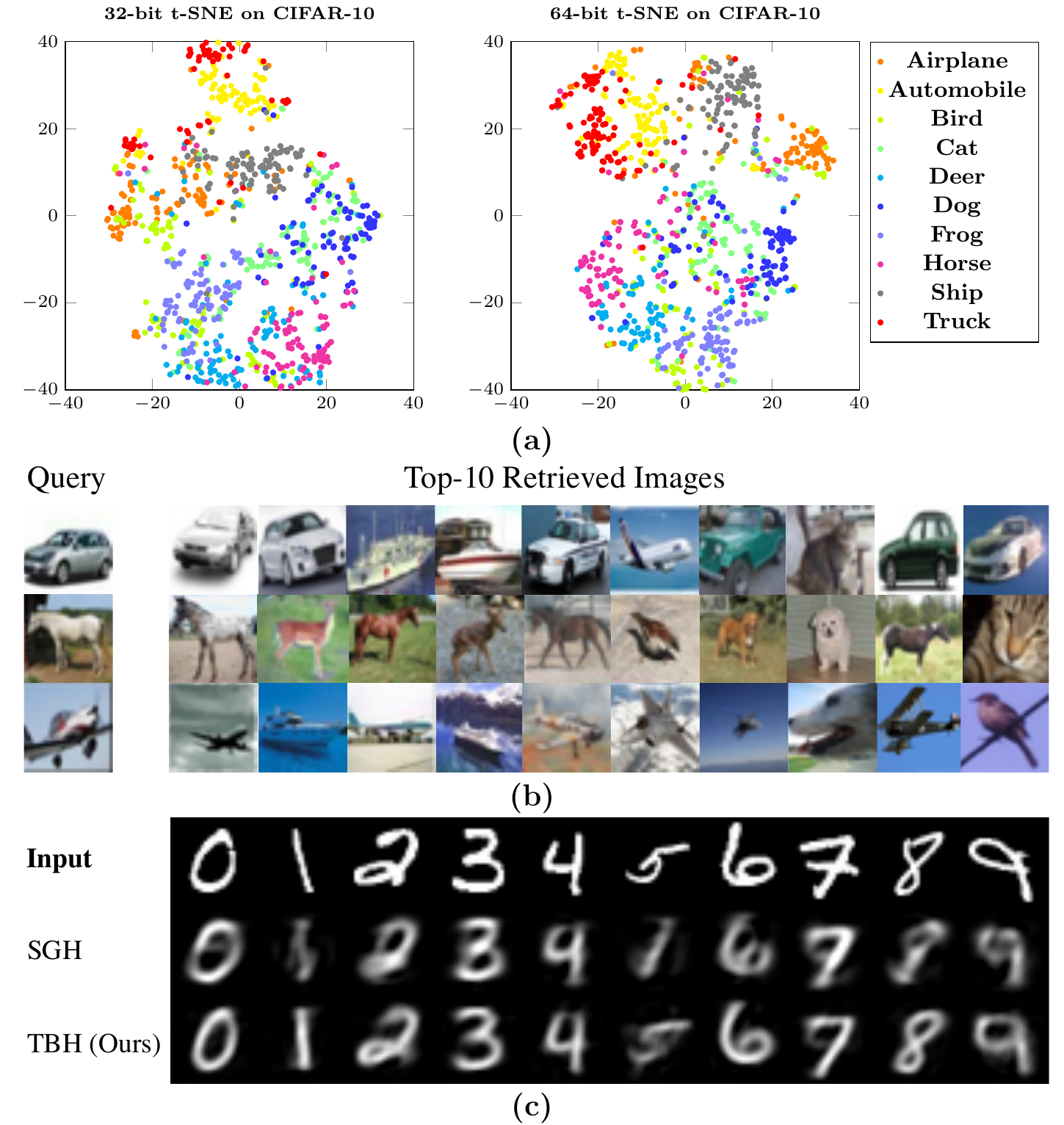}
	\end{center}
	\vspace{-1ex}
	\caption{\textbf{(a)} t-SNE visualization on CIFAR-10. \textbf{(b} 32-bit retrieval results on CIFAR-10. \textbf{(c)} 64-bit image reconstruction results of TBH and SGH on MNIST.}\vspace{-0ex}
	\label{fig_tsne}
\end{figure}

\vspace{-1ex}
Fig.~\ref{fig_tsne} (a) shows the t-SNE~\cite{tsne} visualization results of TBH. Most of the data from different categories are clearly scattered.
Interestingly, TBH successfully locates visually similar categories within short Hamming distances, \eg, \textit{Automobile}/\textit{Truck} and \textit{Deer}/\textit{Horse}. Some qualitative image retrieval results w.r.t. 32-bit codes are shown in Fig.~\ref{fig_tsne} (b).
\vspace{-1ex}\subsubsection{Toy Experiment on MNIST}\vspace{-0ex}
Following \cite{sgh}, another toy experiment on image reconstruction with the MNIST~\cite{mnist} dataset is conducted. For this task, we directly use the flattened image pixels as the network input. The reconstruction results are reported in Fig.~\ref{fig_tsne} (c). %Compared with SGH~\cite{sgh}, TBH decodes images with higher quality. 
Some bad handwritings are falsely decoded to wrong numbers by SGH~\cite{sgh}, while this phenomenon is not frequently observed in TBH. This also supports our insight in addressing the `\textbf{deficient BinBN}' problem. %A single binary bottleneck is generally inadequate in conveying all information required for constructing an information-rich image.% The twin-bottleneck design tackles this problem.%, fully utilizing the auto-encoding mechanism for training unsupervised hashing networks. 
%-------------------------------------------------------------------------

\section{Conclusion}
\vspace{-1ex}In this paper, a novel unsupervised hashing model was proposed with an auto-encoding twin-bottleneck mechanism, namely Twin-Bottleneck Hashing (TBH). The binary bottleneck explored the intrinsic data structure by adaptively constructing the code-driven similarity graph. The continuous bottleneck interactively adopted data relevance information from the binary codes for high-quality decoding and reconstruction. The proposed TBH model was fully trainable with SGD and required no empirical assumption on data similarity. Extensive experiments revealed that TBH remarkably boosted the state-of-the-art unsupervised hashing schemes in image retrieval.

{\small
	\bibliographystyle{ieee_fullname}
	\bibliography{refs}
}

\end{document}